\documentclass[sigconf,nonacm]{acmart}

\usepackage{enumitem}
\usepackage{multirow}
\usepackage{subfigure}
\usepackage{bbm}
\usepackage{amsmath,amsfonts}

\AtBeginDocument{%
  }

\setcopyright{acmlicensed}
\copyrightyear{2025}
\acmYear{2025}
\acmDOI{XXXXXXX.XXXXXXX}

\acmConference[KDD '25]{KDD 2025}{August 03--07,
  2025}{Toronto, Canada}
\acmISBN{978-1-4503-XXXX-X/XX/XX}

\begin{document}

\title{PLMTrajRec: A Scalable and Generalizable Trajectory Recovery Method with Pre-trained Language Models}

\author{Tonglong Wei}
\authornote{Both authors contributed equally to this research.}
\affiliation{%
  \institution{School of Computer and Information Technology, Beijing Jiaotong University}
  \city{Beijing}
  \country{China}
}
\email{{weitonglong}@bjtu.edu.cn}

\author{Yan Lin}
\authornotemark[1]
\author{Youfang Lin}
\affiliation{%
  \institution{School of Computer and Information Technology, Beijing Jiaotong University}
  \city{Beijing}
  \country{China}
}
\email{{ylincs,yflin}@bjtu.edu.cn}

\author{Shengnan Guo}
\authornote{Corresponding author.}
\affiliation{%
  \institution{School of Computer and Information Technology, Beijing Jiaotong University}
  \city{Beijing}
  \country{China}
}
\email{guoshn@bjtu.edu.cn}

\author{Jilin Hu}
\affiliation{%
  \institution{School of Data Science and Engineering, East China Normal University}
  \city{Shanghai}
  \country{China}
}
\email{jlhu@dase.ecnu.edu.cn}

\author{Haitao Yuan}
\author{Gao Cong}
\affiliation{%
  \institution{Nanyang Technological University}
  \country{Singapore}
}
\email{haitao.yuan, gaocong@ntu.edu.sg}

\author{Huaiyu Wan}
\affiliation{%
  \institution{School of Computer and Information Technology, Beijing Jiaotong University}
  \city{Beijing}
  \country{China}
}
\email{hywan@bjtu.edu.cn}

\begin{abstract}
Spatiotemporal trajectory data is crucial for various applications.  
However, issues such as device malfunctions and network instability often cause sparse trajectories, leading to lost detailed movement information. Recovering the missing points in sparse trajectories to restore the detailed information is thus essential.
Despite recent progress, several challenges remain. First, the lack of large-scale dense trajectory data makes it difficult to train a trajectory recovery model from scratch. Second, the varying spatiotemporal correlations in sparse trajectories make it hard to generalize recovery across different sampling intervals. Third, the lack of location information complicates the extraction of road conditions for missing points.

To address these challenges, we propose a novel trajectory recovery model called PLMTrajRec. It leverages the scalability of a pre-trained language model (PLM) and can be fine-tuned with only a limited set of dense trajectories.
To handle different sampling intervals in sparse trajectories, we first convert each trajectory’s sampling interval and movement features into natural language representations, allowing the PLM to recognize its interval. We then introduce a trajectory encoder to unify trajectories of varying intervals into a single interval and capture their spatiotemporal relationships. To obtain road conditions for missing points, we propose an area flow-guided implicit trajectory prompt, which models road conditions by collecting traffic flows in each region. We also introduce a road condition passing mechanism that uses observed points’ road conditions to infer those of the missing points.
Experiments on two public trajectory datasets with three sampling intervals each demonstrate the effectiveness, scalability, and generalization ability of PLMTrajRec.
\end{abstract}




\maketitle

\section{Introduction}
\label{intro}
A trajectory is a sequence of timestamped locations that describe the movement of individuals or vehicles, represented as $\mathcal{T} = \langle p_1,\dots, p_{|\mathcal{T}|} \rangle$ of length $|\mathcal{T}|$, where $p_i = (lat_i, lng_i, t_i)$ denotes the latitude, longitude, and timestamp of the $i$-th point. The sampling interval of $\mathcal{T}$ is $\mu = t_i - t_{i-1}, \forall i \in \{2, \dots, |\mathcal{T}|\}$.
With the advances in GPS devices and geopositioning technologies, a massive amount of trajectory data has been collected. These data play a pivotal role in various applications, such as urban planning~\cite{wang2024cola}, traffic management~\cite{rehena2018towards, wang2021survey}, and personalized location services~\cite{zhao2020go, wang2020next}.
However, factors such as network instability, device malfunctions, or cost-saving settings often lead to missing trajectory points, making the data sparse with large sampling intervals~\cite{ren2021mtrajrec, zhao2019deepmm}. Sparse trajectories fail to accurately reflect movement behavior and route choices, limiting their usefulness. To address this issue, recovering the missing points in sparse trajectories is crucial for preserving trajectory completeness—a task usually referred to as \emph{trajectory recovery}.

Many works have been proposed to tackle trajectory recovery, broadly categorized into \textit{free space trajectory recovery}~\cite{chen2011discovering, chen2023teri} and \textit{map-matched trajectory recovery}~\cite{wei2024micro, chen2023rntrajrec}.
Free space methods directly predict the coordinates of missing points but do not ensure alignment with the road network, typically requiring a subsequent map-matching step. Map-matched methods, by contrast, aim to recover missing points directly on the road network, making them more convenient for downstream applications.
Despite the progress of existing methods, map-matched trajectory recovery still faces the following challenges:

\textbf{1) Limited dense trajectory data.}
Trajectory recovery models are data-driven and typically require large-scale pairs of sparse and dense trajectories. However, because of cost-saving settings and device malfunctions, most collected trajectories remain sparse. The limited availability of dense trajectory data makes existing models prone to overfitting and hampers their performance.

\textbf{2) Difficulty in generalizing to varying sampling intervals.}
Due to network instability and device malfunctions,  a sparse trajectory dataset often contains trajectories with a mixture of different sampling intervals~\cite{wang2021survey}. 
Existing works~\cite{wei2024micro, chen2023rntrajrec, ren2021mtrajrec} usually treat all intervals in the same way, overlooking the varying spatiotemporal correlations of sparse trajectories entangled with multiple sampling intervals~\cite{chen2024multi,che2018hierarchical}.
As a result, when encountering trajectories with unseen sampling intervals, these models often require retraining, which increases costs.

\begin{figure}
  \centering
  \includegraphics[width=\linewidth]{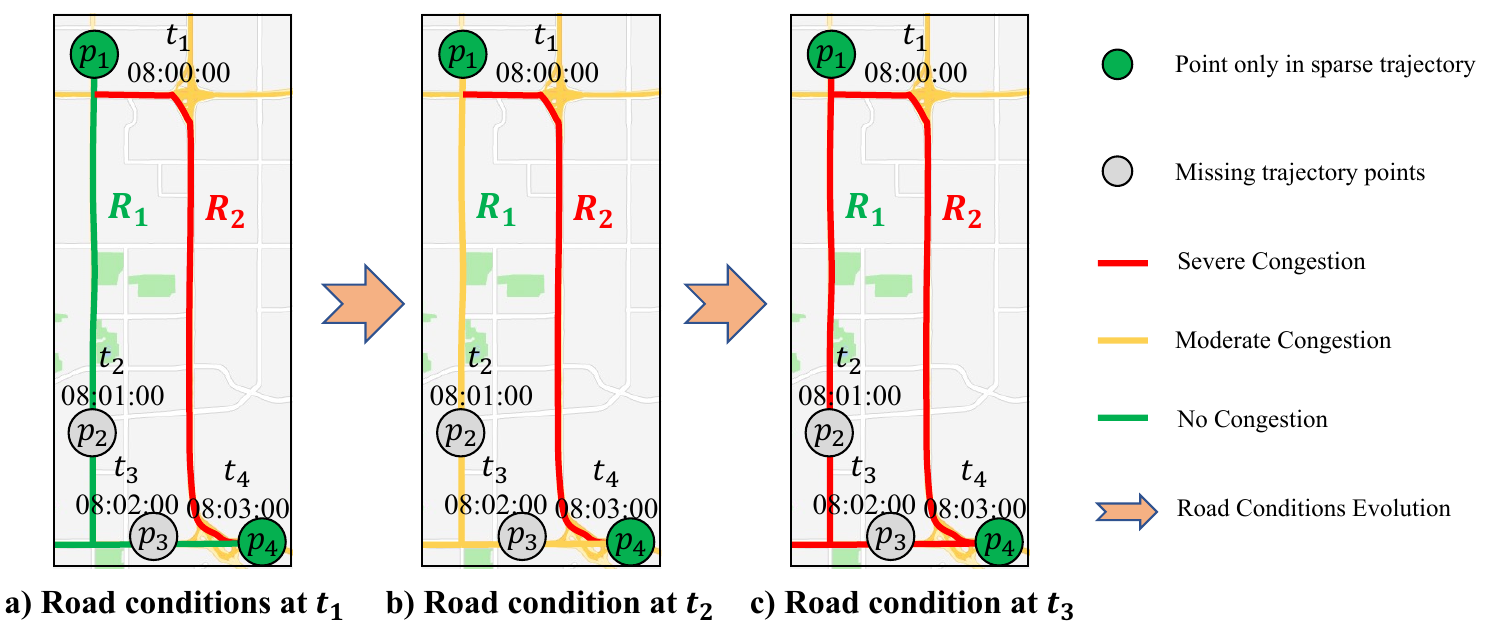}
  \caption{Impact of road conditions on route selection and movement patterns.}
  \label{fig:example}
\end{figure}

\textbf{3) Non-trivial extraction of dynamic road conditions for missing points.}  
Road conditions are key to understanding movement patterns. Existing work~\cite{wei2024micro} incorporates road conditions around observed points but ignores those at missing points. For example, in Figure~\ref{fig:example}(b) and (c), knowing the conditions at $p_2$ and $p_3$ reveals that the user is gradually decelerating on $R_1$ due to congestion. If we only consider conditions at $p_1$ and $p_4$, we might infer that the user takes $R_1$ instead of $R_2$ (Figure~\ref{fig:example}(a)), but not the finer details of the movement. Since the exact locations of missing points are unknown, their road conditions cannot be directly extracted.

To tackle these challenges, we propose \textit{\underline{P}re-trained \underline{L}anguage \underline{M}odel for \underline{Traj}ectory \underline{Rec}overy} (\textbf{PLMTrajRec}). We leverage a pre-trained language model (PLM) that has already learned useful representations from large-scale corpus data and fine-tune it using only a small number of dense trajectories. This enables \mbox{PLMTrajRec} to recover missing points even when dense trajectories are scarce (\textit{Challenge 1}).
To handle sparse trajectories with varying sampling intervals (\textit{Challenge 2}), we introduce an interval and feature-guided (IF-guided) \emph{explicit trajectory prompt}. It encodes both sampling intervals and movement features into the PLM through natural language, helping the model identify these intervals and capture key trajectory characteristics. We then propose an \emph{interval-aware trajectory embedder} to standardize different sampling intervals and learn their spatiotemporal correlations.
To infer road conditions for missing points (\textit{Challenge 3}), we design an area flow-guided (AF-guided) \emph{implicit trajectory prompt} that gathers traffic flows in each region. We also present a \emph{road condition passing mechanism} that uses road conditions from nearby observed points to estimate those of the missing points.

Overall, the main contributions of our work can be summarized as follows:


\begin{itemize}[leftmargin=*]
\item We introduce PLMTrajRec, a trajectory recovery model with strong scalability. By leveraging a pre-trained language model, PLMTrajRec can be fine-tuned on a small set of dense trajectories while still achieving effective recovery.
\item We propose an IF-guided explicit trajectory prompt and an interval-aware trajectory embedder to help PLMTrajRec generalize to sparse trajectories with various sampling intervals. The prompt encodes intervals and movement features into the PLM, and the embedder unifies intervals and learns spatiotemporal correlations.
\item We design an AF-guided implicit trajectory prompt to capture road conditions by aggregating traffic flows, and a road condition passing mechanism to infer missing-point conditions from nearby observations.
\item We conduct extensive experiments on two real-world datasets, each with three sampling intervals, showing that PLMTrajRec achieves state-of-the-art performance in effectiveness, scalability, and generalizability.
\end{itemize}

\section{Related Work}

\textbf{Free-space Trajectory Recovery.}
Trajectory recovery in free space aims to directly restore missing GPS coordinates. Early methods relied on predefined rules to model mobile objects~\cite{chen2016learning, su2013calibrating}, 
such as the vehicle moving with uniform linear motion~\cite{hoteit2014estimating} or taking the most popular route~\cite{chen2011discovering}.
Other approaches~\cite{banerjee2014inferring, zheng2012reducing} use Markov models to capture spatial transitions. However, these methods are limited to modeling low-order transitions and fail to capture the global spatial-temporal dependencies essential for accurate trajectory recovery.
Recent deep learning-based methods have provided more effective solutions~\cite{xu2017trajectory, xu2022temporal}. DHTR~\cite{wang2019deep} employs a GRU-based seq2seq model to analyze user route transitions, integrating Kalman filtering to improve accuracy.
AttnMove~\cite{xia2021attnmove} leverages attention mechanisms and LSTM to capture movement preferences, PeriodicMove~\cite{sun2021periodicmove} models trajectories as graphs using GNNs to extract user patterns. TERI~\cite{chen2023teri} employs a transformer-based two-stage framework for irregular interval trajectory recovery, and TrajBERT~\cite{si2023trajbert} adapts BERT~\cite{devlin2018bert} for implicit trajectory recovery at a coarser granularity.  
Despite their effectiveness, these methods require an additional map-matching step to align recovered trajectories with road networks before they can be applied in navigation systems. This two-stage process leads to error accumulation and significant time overhead due to the computational cost of map-matching.

\textbf{Map-matched Trajectory Recovery.}
Compared with the above methods, map-matched trajectory recovery incorporates the road network as input and aims to recover the trajectory directly onto the road. 
To this end, MTrajRec~\cite{ren2021mtrajrec} first represents the trajectory point with a road segment and moving rate, utilizing the seq2seq-based multi-task framework to capture the spatiotemporal correlations within the trajectory. 
Following this idea, RNTrajRec~\cite{chen2023rntrajrec} models the relationship between trajectories and road networks, designing a spatial-temporal transformer architecture to encode sparse trajectories. 
LightTR~\cite{liu2024lighttr} proposes a lightweight trajectory recovery framework based on federated learning.
MM-STGED~\cite{wei2024micro} models sparse trajectories from a graph perspective, considering both micro and macro semantic information of trajectories.
Although they demonstrate promising progress in trajectory recovery, their performance is still hindered by the poor availability of large-scale dense trajectories and limited in generalization to different sampling intervals discussed in Section~\ref{intro}.

\section{Preliminaries}
\subsection{Definitions}
\begin{definition}
    [Trajectory]
    A trajectory is defined as a series of timestamped locations, denoted as $\mathcal{T} = \langle p_1, \cdots, p_{|\mathcal{T}|} \rangle$ where $p_i = (lat_i, lng_i, t_i)$ represents the latitude and longitude coordinates of an object at the time $t_i, i \in \{1, \cdots, |\mathcal{T}| \}$. $|\mathcal{T}|$ is the length of trajectory. The sampling interval of the trajectory $\mathcal{T}$ is $t_i - t_{i-1}, i \in \{2, \cdots, |\mathcal{T}|\}$.
\label{trajectory}
\end{definition}


\begin{definition}
    [Road Network]
    A road network is defined as a directed graph $\mathcal{G} = (\mathcal{V}, \mathcal{E})$, where $\mathcal{V}$ denotes the set of nodes. Each node $v \in \mathcal{V}$ signifies an intersection that links various road segments, and each node possesses attributes of latitude and longitude. $\mathcal{E}$ represents the set of edges, where each edge $e \in \mathcal{E}$ denotes road segment that connecting two intersections. An edge can be characterized by its starting intersection $e.start \in \mathcal{V}$ and ending intersection $e.end \in \mathcal{V}$.
\end{definition}

\begin{definition}
    [Map-matched Trajectory]
    By employing the map-matching algorithm, a trajectory $\mathcal{T}$ can be projected onto the road network and return the map-matched trajectory $\mathcal{T}_m$. As a result, each point of the map-matched trajectory aligns accurately with a particular road. The  map-matched trajectory is represented as $\mathcal{T}_m = \langle q_1, \cdots, q_{|\mathcal{T}_m|} \rangle$, where each $q_j = (e_j, r_j, t_j)$ signifies the vehicle's location at time $t_j, j \in \{1, \cdots, |\mathcal{T}_m|\}$. Here, $e \in \mathcal{E}$ is the matched road segment, and $r$ is the moving ratio, which quantifies the proportion of the distance traveled from the starting point of a road segment in relation to the total length of that road segment.


    
\end{definition}
\begin{figure}
  \centering
  \includegraphics[width=\linewidth]{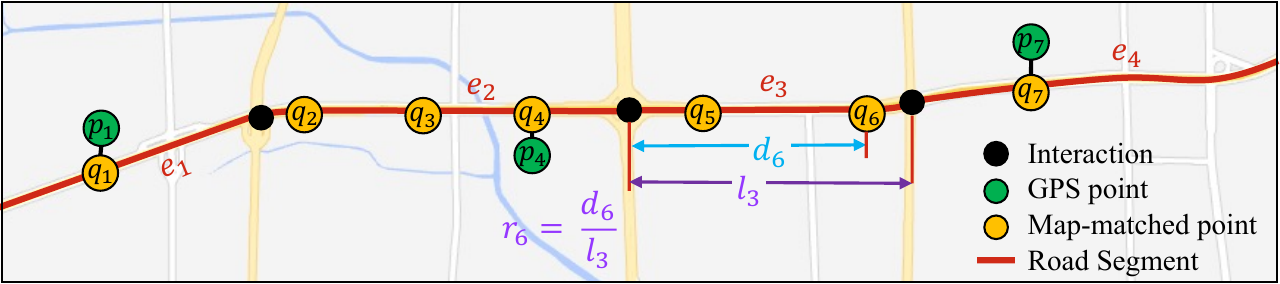}
  \caption{An illustration of map-matched trajectory, road segment $e$, and moving ratio $r$.}
  \label{fig:map_traj}
\end{figure}
\begin{example}
    Figure~\ref{fig:map_traj} gives an example of a map-matched trajectory. For a sparse trajectory $\mathcal{T}_s = \langle p_1, p_4, p_7 \rangle$ with the sampling interval of 90 seconds, its corresponding dense map-matched trajectory is represented by $\mathcal{T}_m = \langle q_1, \cdots, q_7 \rangle$ with a sampling interval of 30 seconds.
    Take a matching point $q_6$ as an example. It falls on the road $e_3$, so its road segment is $e_3$. The distance of $q_6$ from the start point of the road is represented as $d_6$, while the total length of $e_3$ is denoted by $l_3$. 
    Therefore, the moving ratio $r_6$ can be calculated as $r_6 = \frac{d_6}{l_3}$.
\end{example}

\begin{figure*}
  \centering
  \includegraphics[width=\linewidth]{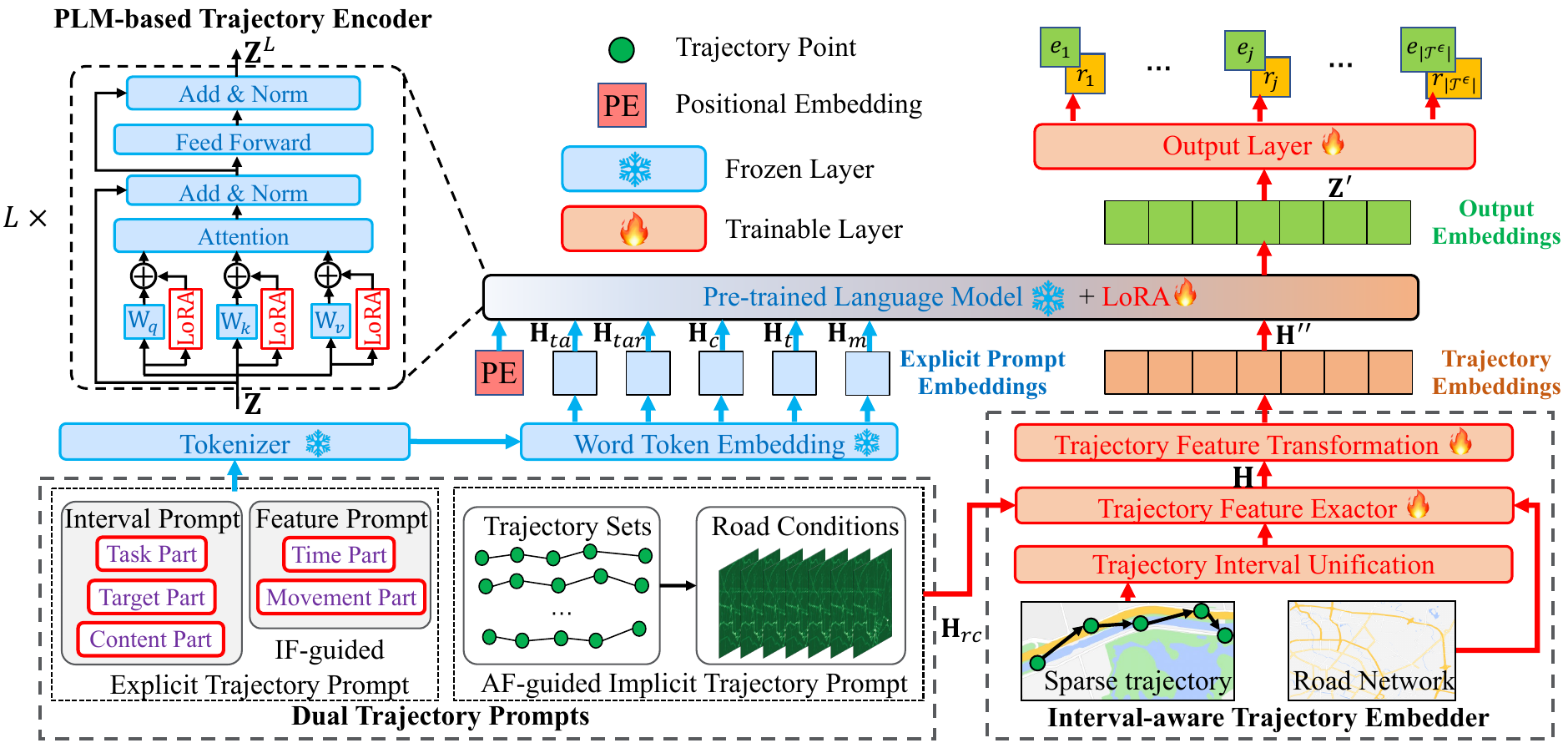}
  \caption{The framework of PLMTrajRec, consists of Dual Trajectory Prompts, Interval-aware Trajectory Embedder, and PLM-based Trajectory Encoder.}
  \label{fig:model}
\end{figure*}

\subsection{Problem Statement}
\textbf{Map-matched Trajectory Recovery}.
Given a sparse trajectory $\mathcal{T}_s = \langle p_1, \cdots, p_{|\mathcal{T}_s|} \rangle$ with a sampling interval of $\mu$. The goal of map-matched trajectory recovery is to reconstruct the dense map-matched trajectory $\mathcal{T}_m = \langle q_1, \cdots, q_{|\mathcal{T}_m|} \rangle$ with a sampling interval of $\epsilon$. Note that the sampling interval $\mu$ is larger than $\epsilon$.

\section{Methodology}

\subsection{Overall Pipeline}
In this paper, we present both scalable and generalizable trajectory recovery model, PLMTrajRec, by fine-tuning a PLM that is pre-trained on a large-scale corpus with limited dense trajectory data. The framework of PLMTrajRec, shown in Figure~\ref{fig:model}, comprises three main components: dual trajectory prompts, an interval-aware trajectory embedder, and a PLM-based trajectory encoder.

The dual trajectory prompts provide essential information through two key components. 
First, the interval and feature-guided (IF-guided) explicit trajectory prompt, expressed in natural language, injects the sampling interval of sparse trajectories and their movement features into PLM. This enhances the model’s ability to recognize the sampling intervals of each sparse trajectory and capture trajectory characteristics.
Second, area flow-guided (AF-guided) implicit trajectory prompts model road conditions, offering crucial contextual information to improve the recovery of missing trajectory points.

The interval-aware trajectory embedder first normalizes sparse trajectories with varying sampling intervals $\mu$ into a unified interval $\epsilon$, effectively handling the diverse spatiotemporal correlations associated with different sampling intervals and enhancing model generalization. Subsequently, each trajectory point, including both observed and missing points, is encoded into an embedding that the PLM can process.

In the PLM-based trajectory encoder, pre-trained BERT is employed to capture information bi-directionly, and most of the parameters are frozen to preserve the pre-trained knowledge, while the multi-head attention layer remains trainable, allowing the model to capture the spatiotemporal correlations within the trajectory. Finally, the output layer predicts the road segment $e$ and the moving ratio $r$ for the recovered map-matched trajectory at each step.

\subsection{Dual Trajectory Prompts}
To enable PLM to recover trajectories with different intervals and effectively model road conditions for missing trajectory points, we introduce dual trajectory prompts, including an interval and feature-guided (IF-guided) explicit trajectory prompt and an area flow-guided (AF-guided) implicit trajectory prompt.

\subsubsection{IF-guided Explicit Trajectory Prompt}
The IF-guided explicit trajectory prompt offers a structured textual description of the sampling intervals and movement features of sparse trajectories, enabling PLMTrajRec to identify varying sampling intervals and capture trajectory features. 
The prompts related to the sampling interval include the following components:

\begin{itemize}[leftmargin=*]
    \item \textbf{<Task Part>}: Sparse trajectory recovery. 
    \item \textbf{<Target Part>}: Output the road segment and moving ratio for each point in the trajectory. 
    \item \textbf{<Content Part>}: The sparse trajectory is sampled every $\mathit{\{large}$  $\mathit{sampling}$ $\mathit{interval\}}$  and aims to recover trajectory every $\mathit{\{small}$  $\mathit{sampling}$ $\mathit{interval\}}$  seconds. 
\end{itemize}
The content within the placeholders  $\{\}$ is filled with trajectory-specific information. 
All trajectories share the same interval-related prompt. The <Task Part> informs the PLM about the overall task to be performed, the <Target Part> defines the required output format, and the <Content Part> specifies the sampling intervals, guiding the PLM in effectively analyzing the trajectories.

The prompts related to the movement features include the following contents:
\begin{itemize}[leftmargin=*]
     
     \item \textbf{<Time Part>}: The trajectory started at \{$\mathit{start}$ $\mathit{time}$\} on $\mathit{\{day}$-$\mathit{in}$-$\mathit{week\}}$ and ended at \{$\mathit{end}$ $\mathit{time}$\} on $\mathit{\{day}$-$\mathit{in}$-$\mathit{week\}}$.
    \item \textbf{<Movement Part>}: Total time cost: \{$\mathit{x}$ 
 $\mathit{minutes}$ $\mathit{y}$ $\mathit{seconds}$\}.
    Total space transfer distance: \{$\mathit{z}$  $\mathit{kilometers}$\}.
 \end{itemize}

The <Time Part> provides the trajectory's specific start and end times, helping the PLM understand the duration and potential time patterns, such as morning or evening peaks. The <Movement Part> supports the PLM in inferring the trajectory’s movement. We give a detailed example of the IF-guided explicit trajectory prompt in Appendix~\ref{trajectory_prompt_example}.

After obtaining the prompt for each part, we use the tokenizer and word token embedding in PLMs to convert text into embeddings. The embedding of <Task Part>, <Target Part>, <Content Part>, <Time Part>, and <Movement Part> are denoted by $\textbf{H}_{ta}$, $\textbf{H}_{tar}$, $\textbf{H}_{c}$, $\textbf{H}_{t}$, and $\textbf{H}_{m}$.
These embeddings are then concatenated to form the overall IF-guided explicit trajectory prompt embedding.
\begin{equation}
    \textbf{H}^{e} = \textbf{H}_{ta} || \textbf{H}_{tar} || \textbf{H}_{c} || \textbf{H}_{t} || \textbf{H}_{m},
\end{equation}
where $ ||  $ is the concatenate operation.


\subsubsection{AF-guided Implicit Trajectory Prompt}
\label{implicit_prompt}
Road conditions can provide insights into both the surrounding environment and the object's movement, which are helpful for trajectory recovery. 
For instance, vehicles typically slow down in congested areas and accelerate in less congested regions. Given the complexity and variability of real-world road conditions are difficult to express in natural language, we represent them as the implicit trajectory prompt.

To obtain the road conditions for each trajectory point, we first calculate the average road conditions across all areas and time intervals, then extract the relevant information for each point. Specifically, we divide the area of interest into a spatial grid of $ I \times J $ cells and partition the time into $T$ slices. For each grid cell, we compute the total traffic, resulting in a regional flow matrix $\textbf{RC} \in \mathbb{R}^{I \times J \times T}$. Each element in $\textbf{RC}$ represents the number of traffic flows at region $i,j$ at time $t$.
To capture the spatiotemporal correlations among different regions, we apply a 2D convolution (2D CNN) along the spatial dimension and a 1D CNN along the temporal dimension, yielding the road condition representation $\textbf{H}_{rc} \in \mathbb{R}^{I \times J \times T \times F}$, where $F$ is the number of features. Formally:
\begin{equation}
    \mathbf{H}_{rc} = \text{Conv1d(Conv2d(\textbf{RC}))} \in \mathbb{R}^{I \times J \times T \times F}
\end{equation}

Obviously, extracting the road conditions of a trajectory point needs its longitude, latitude, and timestamp.  
However, it is challenging for missing points as their geographic coordinates are unknown. 
Inspired by the message passing mechanism~\cite{gilmer2017neural}, we introduce a \textit{road condition passing mechanism} that approximates the road conditions of missing points by leveraging the surrounding road conditions. 
The implementation details will be elaborated in Section~\ref{feature_extract}.

\begin{figure}
  \centering
  \includegraphics[width=\linewidth]{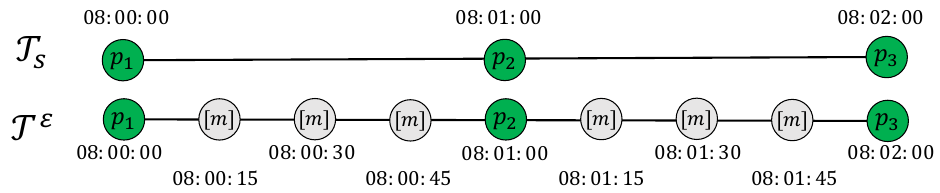}
  \caption{An illustration of trajectory preprocessing layer.}
  \label{fig:encoder}
\end{figure}

\subsection{Interval-aware Trajectory Embedder}
To make the model deal with sparse trajectories with different sampling intervals and capture the spatiotemporal correlation, we propose an interval-aware trajectory embedder.

\subsubsection{Trajectory Interval Unification}
For different sparse trajectories, their sampling intervals $\mu$ are not constant, such as 4 minutes, 2 minutes, or 1 minute, which bring different spatiotemporal correlations between trajectory points. To this end, we convert these into the same sampling interval with the target trajectory by introducing a placeholder `[m]' to mark missing trajectory points and generate a preprocessed sparse trajectory $\mathcal{T}^{\epsilon}$, where the length of $\mathcal{T}^{\epsilon}$ is $\frac{p_{|\mathcal{T}_s |}.t - p_1.t}{\epsilon } + 1$.
Although the location of `[m]' is unknown, its timestamp can still be calculated, as illustrated in Figure~\ref{fig:encoder}.

\subsubsection{Trajectory Feature Extractor}
\label{feature_extract}
Given the preprocessed sparse trajectory $\mathcal{T}^{\epsilon}$, there are two cases for extract trajectory point features: 

\begin{itemize}[leftmargin=*]
     \item Case 1: the trajectory point $s \in \mathcal{T}^{\epsilon}$ is observed. i.e., $\exists \, k\in\{1, \cdots , |\mathcal{T}_s |\}, p_k.t = s.t$.
    \item Case 2: the location of trajectory point $s \in \mathcal{T}^{\epsilon}$ is missing. i.e., $s = [m]$, note that its time $s.t$ is known.
\end{itemize}

For case 1, we use Learnable Fourier Features (LFF)~\cite{tancik2020fourier, li2021learnable} to encode the continuous latitude and longitude of $s$ into an $F$-dimensional vector using the feature mapping function $\Phi(x): \mathbb{R} \to \mathbb{R}^F$. Formally:
\begin{equation}
    \Phi (x) = W_{\Phi } [\mathrm{cos} \text{ }x W_r || \mathrm{sin} \text{ }x W_r],
\end{equation}
where $W_r \in \mathbb{R}^{F/2}$ and $W_{\Phi } \in \mathbb{R}^{F \times F}$ represent the learnable parameters, and $x \in \{s.lat, s.lon\}$.
Using the feature mapping function $\Phi(\cdot)$, the relative information $x - y$ between points $x$ and $y$ can be captured through multiplication operations, which is important for understanding movement, such as distance. We provide detailed insight and proof of LFF in Appendix~\ref{insight_LLF}.

Additionally, considering that vehicles move within the road network, the relationship between a trajectory point and its surrounding road segments is critical. To measure the relationship between a trajectory point $s$ and a road segment $l$, we define a function $f(d_{s,l})$ based on their shortest distance:
\begin{equation}
\label{cal_road_weight}
    f(d_{s,l}) = \begin{cases}
  e^{-(\frac{d_{s,l}}{\kappa } )^2}& \text{ if } d_{s,l} < \varphi_{dist}, \\
  0 & \text{ otherwise ,}
\end{cases}
\end{equation}
where $d_{s,l}$ is the shortest distance between $s$ and road $l$, $\kappa$ is a hyperparameter, and $\varphi_{dist}$ is a distance threshold. 
Then, we obtain the road network representation $\mathbf{h}_s^{\text{road}}$ by:
\begin{equation}
    \mathbf{h}_s^{\text{road}} = \frac{\sum_{l=1}^{|\mathcal{V} |}f(d_{s,l}) \cdot \textbf{M}_l}{\sum_{l=1}^{|\mathcal{V} |}f(d_{s,l})},
\end{equation}
where $\textbf{M}_l \in \mathbb{R}^F$ is the learnable embedding of road segment $l$, and $\mathcal{|V|}$ is the total number of road segments.
Finally, the complete representation of a trajectory point $s$ is obtained as:
\begin{equation}
    \mathbf{h}_s = W_1[(\Phi(s.lat) + \Phi(s.lng) ) || \mathbf{h}_s^{\text{road}}] + b_1,
\end{equation}
where $W_1\in \mathbb{R}^{F \times 2F}$ and $b_1 \in \mathbb{R}^F$ is the learnable parameters.

For case 2, where the location of point $s$ is unknown, here we use road conditions to represent its features, as described in Section~\ref{implicit_prompt}. 
Since the road conditions at a specific point are influenced by the surrounding conditions, which propagate along both temporal and spatial dimensions, we propose a road condition passing mechanism to infer the road conditions of the missing location inspired by the message passing mechanism~\cite{gilmer2017neural}.

\textbf{Road Condition Passing Mechanism.} First, for the missing point $s$, we identify the observed forward and backward trajectory points, $s_f$ and $s_b$, respectively. 
For instance, in Figure~\ref{fig:encoder}, suppose we have a missing point $s$ timestamped at 8:00:45, its observed forward point $s_f = p_1$ and the backward point $s_b = p_2$. Then, we retrieve the road conditions of $s_f$ and $s_b$ as follows:
\begin{equation}
\begin{split}
    \mathbf{h}_{rc}^f &= \mathbf{H}_{rc}[\pi _{lat}(s_f.lat), \pi _{lng}(s_f.lng), \pi _{t}(s_f.t)], \\
    \mathbf{h}_{rc}^b &= \mathbf{H}_{rc}[\pi _{lat}(s_b.lat), \pi _{lng}(s_b.lng), \pi _{t}(s_b.t)],
\end{split}
\end{equation}
where $\pi_{lat}$, $\pi_{lng}$, and $\pi_{t}$ are index functions mapping latitude, longitude, and time to their corresponding indices.
Next, we calculate the time interval between $s$ and $s_f$ and $s_b$, denoted as $\triangle t_f = s.t - s_f.t$ and $\triangle t_b = s_b.t - s.t$. The road condition of point $s$ is then computed as:
\begin{equation}
    \mathbf{h}_{rc}^s = \frac{e^{-\triangle t_f}\mathbf{h}_{rc}^f + e^{-\triangle t_b}\mathbf{h}_{rc}^b}{e^{-\triangle t_f} + e^{-\triangle t_b}} 
\end{equation}

In addition, we encode the time intervals $\triangle t_f$ and $\triangle t_b$ to quantify the relative position of $s$. 
The final representation of $s$ is given by:
\begin{equation}
    \mathbf{h}_s = W_2 [ \textbf{m} \text{ } || \text{ FC}(\triangle t_f || \triangle t_b) \text{ } || \text{ } \mathbf{h}_{rc}^s] + b_2,
\end{equation}
where $\textbf{m} \in \mathbb{R}^{F}$ is a learnable vector to represent the location is missing, and $\text{FC}(\cdot): \mathbb{R}^2 \to \mathbb{R}^F$ is the fully connection layer. $W_2 \in \mathbb{R}^{F \times 3F}$ and $b_2 \in \mathbb{R}^F$ are the learnable parameters.
Finally, the overall trajectory representation is $\textbf{H} \in \mathbb{R}^{|\mathcal{T}^{\epsilon}| \times F}$.

\subsubsection{Trajectory Feature Transformation}
To enhance the model's ability to comprehend trajectory features, we introduce $K$ reference tokens $\textbf{E}_w \in \mathbb{R}^{K \times F}$, inspired by~\cite{lee2019set, guo2023self}, to bridge the connection between PLMs and the trajectory. These reference tokens are designed to capture the global semantics of the trajectory.

Given the trajectory embedding $\textbf{H}$, we first apply a 1D CNN to aggregate neighboring information and capture local movement patterns:
\begin{equation}
    \textbf{H}' = \text{Conv1d}(\textbf{H})
\end{equation}

Next, we compute self-attention between the trajectory embedding $\textbf{H}'$ and the reference tokens $\textbf{E}_w$ to capture the global semantics, where $\textbf{H}'$ acts as the query, and $\textbf{E}_w$ as both the key and value:
\begin{equation}
    \textbf{H}'' = \text{Attention}(\textbf{H}', \textbf{E}_w, \textbf{E}_w)
\end{equation}

After transforming the embedding $\textbf{H}''$ in the token space, we concatenate it with the explicit trajectory prompt embedding $\textbf{H}^e$ to form the final trajectory representation:
\begin{equation}
    \textbf{Z} = \textbf{H}^e || \textbf{H}''
\end{equation}
Then, we add the Transformer positional embedding $PE$ in each element of $\textbf{Z}$ and feed it into PLMs encoder to encode trajectory.

\subsection{PLM-based Trajectory Encoder}
We use the pre-trained BERT as the fundamental architecture of PLMs, considering that its encoder-only structure is well-suited to the reconstruction task due to its effective use of bi-directional contextual information from the trajectory~\cite{sun2023test}. To make PLM fully adaptable to the trajectory recovery task, we employ the Low-Rank Adaptation (LoRA) algorithm~\cite{hu2021lora}. LoRA introduces additional parameters within the Transformer block, reducing the number of trainable parameters and computational complexity while enhancing the PLM’s performance on new tasks.

Specifically, in each self-attention block of the Transformer, we introduce additional weights $\Delta \textbf{W}_q \in \mathbb{R}^{F \times F}$, $\Delta \textbf{W}_k \in \mathbb{R}^{F \times F}$, and $\Delta \textbf{W}_v \in \mathbb{R}^{F \times F}$ for the query, key, and value matrices $\textbf{W}_q$, $\textbf{W}_k$, and $\textbf{W}_v$. To reduce the parameter count, the additional weights $\Delta \textbf{W}$ are decomposed into two low-rank matrices, $\textbf{B} \in \mathbb{R}^{F \times r}$ and $\textbf{C} \in \mathbb{R}^{r \times F}$, where $r \ll F$. The modified query, key, and value matrices in each self-attention block are updated as $\textbf{W}_q = \textbf{W}_q + \Delta \textbf{W}_q$, $\textbf{W}_k = \textbf{W}_k + \Delta \textbf{W}_k$, and $\textbf{W}_v = \textbf{W}_v + \Delta \textbf{W}_v$.
Formally, the output of the $l$-th transformer layer is defined as:
\begin{equation}
    \textbf{Z}^l = \text{LoRA(}\text{Transformer }(\textbf{Z}^{l-1}))
\end{equation}

After stacking $L$ Transformer layers, we discard the IF-guided explicit trajectory prompt portion and obtain the output embeddings $\mathbf{Z}' \in \mathbb{R}^{|\mathcal{T}^{\epsilon} | \times d}$.
For the $j$-th element of $\textbf{Z}'$, we apply the softmax function to calculate the probability of each road segment and use the argmax function to return the predicted road segment $e_j$. We employ a MLP with a Sigmoid activation function to determine the moving ratio $r_j$.
\subsection{Training}
\subsubsection{Loss Function}
We use multi-task learning to optimize both road segment recovery and moving ratio recovery simultaneously.
For the road segment, we use the cross-entropy loss function:
\begin{equation}
    \mathcal{L}_e = -\frac{1}{|\mathcal{T}^{\epsilon} |}\sum_{j=1}^{|\mathcal{T}^{\epsilon} |}\mathrm{log}(\hat{e}_j|\textbf{Z}'_j  )    
\end{equation}
For the moving ratio, we use the mean squared error loss function:
\begin{equation}
    \mathcal{L}_r = \frac{1}{|\mathcal{T}^{\epsilon} |}\sum_{j=1}^{|\mathcal{T}^{\epsilon} |}|r_j - \hat{r}_j |^2
\end{equation}

The final loss function is $\mathcal{L} = \mathcal{L}_e + \lambda \mathcal{L}_r$, where $\lambda$ is a hyper-parameter to balance two tasks.

\subsubsection{Joint Training Strategy}
To generalize PLMTrajRec in recovering sparse trajectories with various sampling intervals, we propose the following training approach.

For the dense trajectory dataset $\mathbb{T} $, which is set as a 15-second sampling interval in our experiments.
To create sparse trajectory datasets, we re-sample each trajectory $\mathcal{T} \in \mathbb{T}$ at three different intervals: 1 minute, 2 minutes, and 4 minutes, resulting in three datasets: $\mathbb{T}_1$, $\mathbb{T}_2$, and $\mathbb{T}_4$. These are then combined to form a larger training dataset $\mathbb{T}_{all}$. Formally,
\begin{equation}
    \mathbb{T} _{all} = \mathbb{T}_1\cup \mathbb{T}_2\cup \mathbb{T}_4.
\end{equation}

By training PLMTrajRec on $\mathbb{T}_{all}$, we achieve balanced performance across the different sampling intervals. Subsequently, we fine-tune the model on each individual sparse trajectory dataset to further improve trajectory recovery performance.

\section{Experiments}
In this section, we present comprehensive experiments to evaluate the effectiveness, scalability, and generalization capabilities of PLMTrajRec on two real-world trajectory datasets with three different sparse trajectory sampling intervals. 
    


\subsection{Datasets}

We assess our model using two publicly available trajectory datasets from Chengdu, China, and Porto, Portugal. The Chengdu dataset\footnote{\url{https://outreach.didichuxing.com/}} contains trajectory data collected in November 2016. The Porto dataset\footnote{\url{https://www.kaggle.com/competitions/pkdd-15-predict-taxi-service-trajectory-i/data}} includes trajectory from 442 taxis, collected from July 2013 to June 2014. Both datasets are standardized to a sampling interval of 15 seconds. We remove trajectories with travel times less than 5 minutes or exceeding 1 hour, as well as outlier trajectories. The road network data is obtained from the OpenStreetMap website\footnote{\url{http://www.openstreetmap.org/}}. We employ a map-matching algorithm~\cite{newson2009hidden} to project the trajectory onto the road network and obtain the ground truth of the road segments and moving rates. The detailed overview of the dataset characteristics is provided in Table~\ref{tab:dataset}.

\begin{table}[t]	
\centering
\caption{Dataset Description.}
\begin{tabular}{c|c|c}
    \toprule
    Types & Chengdu & Porto \\
    \hline
    Sampling interval & 15s & 15s \\
    \# Trajectory & 118,354  & 322,079 \\
    \# Road segment & 2504 & 2224 \\
    Latitude range & 30.655 $\sim$ 30.727 & 41.142 $\sim$ 41.174 \\
    Longitude range & 104.043 $\sim$ 104.129 & -8.652 $\sim$ -8.578 \\
    Length / Width & 8.22 km / 8.00 km & 6.19 km / 3.56 km \\
    \bottomrule
\end{tabular}
\label{tab:dataset}
\end{table}

\subsection{Baselines}
To evaluate the effectiveness of our model, we compare PLMTrajRec with 12 baseline methods. These include five free space trajectory recovery models: \textbf{HMM~\cite{newson2009hidden} + ShortestPath}, \textbf{Linear~\cite{hoteit2014estimating} + HMM~\cite{newson2009hidden}}, \textbf{MPR~\cite{chen2011discovering} + HMM~\cite{newson2009hidden}}, \textbf{DHTR~\cite{wang2019deep} + HMM~\cite{newson2009hidden}}, and \textbf{AttnMove~\cite{xia2021attnmove} + Rule}, as well as seven map-matched trajectory recovery models: \textbf{MTrajRec~\cite{ren2021mtrajrec}}, \textbf{T2vec~\cite{li2018deep} + Decoder}, \textbf{T3s~\cite{yang2021t3s} + Decoder}, \textbf{TERI~\cite{chen2023teri} + Decoder}, \textbf{TrajBERT~\cite{si2023trajbert} + Decoder}, \textbf{RNTrajRec~\cite{chen2023rntrajrec}}, and \textbf{MM-STGED~\cite{wei2024micro}}. Due to space constraints, the details of each baseline are provided in Appendix~\ref{append_baseline}.

\subsection{Evaluation Metrics}
We employ five common metrics to assess the effectiveness of our model followed~\cite{ren2021mtrajrec, chen2023rntrajrec, wei2024micro}. For road segment recovery, we utilize \textbf{Accuracy (Acc)}, \textbf{Recall}, and \textbf{Precision (Prec)} metrics to evaluate. 
And employ \textbf{Mean Absolute Error (MAE)} and the \textbf{Root Mean Square Error (RMSE)} in assessing the recovered GPS coordinates. 
The details of each metric are provided in Appendix~\ref{append_metric}.

\begin{table*}[t!]	
\small
\centering
\caption{Performance comparison on two datasets with sampling intervals at 4 minutes, 2 minutes, and 1 minute, respectively.\\
The best results are highlighted in \textbf{bold}, while the \underline{underline} indicates the second-best results.}
\renewcommand{\arraystretch}{0.6}
\resizebox{1.0\linewidth}{!}{
\begin{tabular}{c|c|ccccc|ccccc}
    \toprule
    \multirow{2}{*}{Sampling Interval}&\multirow{2}{*}{Methods}&\multicolumn{5}{c|}{Chengdu} &\multicolumn{5}{c}{Porto}\\
    \cline{3-12}
    & & Acc($\%$) & Recall($\%$) & Prec($\%$) & MAE & RMSE & Acc($\%$) & Recall($\%$) & Prec($\%$) & MAE & RMSE\\
    \midrule
    \multirow{16}{*}{\parbox{2.5cm}{\centering $\mu = 4$ minutes $\to \epsilon = 15$ seconds}} & HMM + ShortestPath & 26.85 & 28.64 & 29.55 & 939.3 & 1047.7 & 20.19 & 26.22 & 33.51 & 886.9 & 941.5 \\
    & Linear + HMM & 26.42 & 30.45 & 36.15 & 974.5 & 1145.4 & 32.23 & 36.09 & 49.80 & 489.3 & 637.3\\
    & MPR + HMM & 36.93 & 38.62 & 44.53 & 821.9 & 914.1 & 32.22 & 38.67 & 48.07 & 534.0 & 700.2\\
    & DHTR + HMM & 41.48 & 57.34 & 50.48 & 673.6 & 911.3 & 32.02 & 58.29 & 45.61 & 456.7 & 627.1 \\
    & AttnMove + Rule & 63.43 & 73.97 & 78.72 & 358.2 & 916.7 & 49.31 & 48.62 & 78.03 & 310.0 & 621.3\\
    & MTrajRec & 65.79 & 75.14 & 78.42 & 315.1 & 904.4 & 52.36 & \underline{60.39} & 77.28 & 266.1 & 590.1\\
    & T3s + Decoder & 65.60 & 75.26 & 78.14 & 318.2 & 926.3 & 52.24 & 60.24 & 77.80 & 270.4 & 594.9 \\
    & T2vec + Decoder & 66.51 & 75.68 & 78.27 & 307.5 & 915.2 & 53.13 & 60.27 & 77.62 & 256.3 & 571.0\\
    & TERI + Decoder & 66.42 & 75.59 & 78.36 & 309.2 & 903.8 & 53.59 & 60.02 & 78.26 & 253.9 & 558.3 \\
    & TrajBERT + Decoder & 66.09 & 75.38 & 78.59 & 310.7 & 911.4 & 52.98 & 60.12 & 77.93 & 251.8 & 560.1 \\
    & RNTrajRec & 67.66 & 75.59 & 79.97 & 306.1 & 886.0 & 54.59 & \textbf{60.42} & 79.20 & 248.1 & 549.1\\
    & MM-STGED & 70.64 & 76.04 & 81.63 & 266.2 & 829.7 & 57.30 & 59.48 & 80.21 & 222.8 & 510.4\\
    \cmidrule{2-12}
    & \textbf{PLMTrajRec} & \underline{74.12} & \underline{79.63} & \underline{86.46} & \underline{262.8} & \underline{483.0} & \underline{57.61} & {59.15} & \textbf{82.19} & \underline{200.9} & \underline{376.9}\\ 
    & \textbf{PLMTrajRec + FT} & \textbf{74.58} & \textbf{80.09} & \textbf{86.63} & \textbf{253.2} & \textbf{465.7} & \textbf{57.67} & {59.08} & \underline{82.05} & \textbf{200.8} & \textbf{370.2}\\
    \midrule
    \multirow{16}{*}{\parbox{2.5cm}{\centering $\mu = 2$ minutes $\to \epsilon = 15$ seconds}} & HMM + ShortestPath & 33.85 & 47.89 & 48.31 & 754.1 & 826.2 & 27.30 & 40.05 & 46.00 & 647.3 & 747.0\\
    & Linear + HMM & 43.78 & 45.35 & 48.77 & 816.9 & 1054.7 & 49.35 & 50.45 & 63.87 & 408.7 & 609.8\\
    & MPR + HMM & 49.88 & 54.62 & 50.94 & 474.8 & 899.0 & 49.76 & 52.97 & 62.86 & 409.8 & 610.9\\
    & DHTR + HMM & 47.17 & 60.16 & 51.73 & 662.0 & 912.2 & 43.76 & 65.25 & 52.10 & 385.3 & 578.0\\
    & AttnMove + Rule & 71.98 & 77.42 & 80.67 & 291.1 & 764.8 & 61.39 & 60.90 & 82.98 & 213.4 & 468.5\\
    & MTrajRec & 74.52 & 78.25 & 81.09 & 254.5 & 885.7 & 61.65 & 65.65 & 78.99 & 179.9 & 451.5\\
    & T3s + Decoder & 74.62 & 78.95 & 81.79 & 242.2 & 857.5 & 61.75 & 65.53 & 79.14 & 181.3 & 461.2 \\
    & T2vec + Decoder & 75.69 & 78.86 & 81.68 & 231.6 & 783.6 & 62.24 & 65.77 & 78.97 & 173.9 & 438.0\\
    & TERI + Decoder & 75.32 & 78.74 & 81.38 & 239.0 & 823.7 & 62.14 & 65.39 & 79.18 & 178.3 & 443.9 \\
    & TrajBERT + Decoder & 75.20 & 78.69 & 81.53 & 235.0 & 813.7 & 62.34 & 65.66 & 79.02 & 177.9 & 441.2 \\
    & RNTrajRec & 75.80 & 79.35 & 81.86 & 218.5 & 757.0 & 63.39 & 65.84 & 79.25 & 171.3 & 433.9 \\
    & MM-STGED & 78.14 & 80.06 & 83.58 & 197.2 & 696.0 & 65.69 & 66.15 & 80.74 & 152.5 & 400.8\\
    \cmidrule{2-12}
    & \textbf{PLMTrajRec} & \underline{81.76} & \underline{84.31} & \underline{88.38} & \underline{187.4} & \underline{366.2} & \underline{66.40} & \underline{66.52} & \underline{82.14} & \underline{141.9} & \underline{294.6}\\ 
    & \textbf{PLMTrajRec + FT} & \textbf{82.29} & \textbf{84.59} & \textbf{88.72} & \textbf{181.2} & \textbf{349.0} & \textbf{66.72} & \textbf{66.87} & \textbf{82.38} & \textbf{141.7} & \textbf{293.5}\\

    \midrule
    \multirow{16}{*}{\parbox{2.5cm}{\centering $\mu = 1$ minute $\to \epsilon = 15$ seconds}} & HMM + ShortestPath & 35.92 & 67.92 & 60.16 & 529.7 & 638.2 & 34.75 & 48.46 & 48.67 & 527.2 & 659.3\\
    & Linear + HMM & 68.59 & 65.66 & 66.67 & 707.4 & 1005.2 & 66.17 & 64.72 & 75.22 & 368.3 & 571.1\\
    & MPR + HMM & 62.25 & 62.67 & 60.53 & 418.8 & 659.0 & 66.27 & 65.66 & 74.51 & 402.6 & 628.2\\
    & DHTR + HMM & 51.09 & 63.40 & 50.14 & 584.7 & 750.4 & 52.98 & 69.60 & 57.17 & 420.4 & 625.8 \\
    & AttnMove + Rule & 79.60 & 81.55 & 82.75 & 194.4 & 752.6 & 72.07 & 69.59 & 80.41 & 156.6 & 360.7 \\
    & MTrajRec & 81.12 & 81.73 & 83.75 & 187.1 & 718.4 & 71.65 & 70.92 & 80.35 & 114.9 & 332.3 \\
    & T3s + Decoder & 80.90 & 82.78 & 83.15 & 187.1 & 713.0 & 71.78 & 71.61 & 80.16 & 110.1 & 328.0\\
    & T2vec + Decoder & 81.69 & 81.90 & 83.88 & 185.6 & 714.1 & 71.86 & 71.10 & 80.48 & 114.2 & 334.7 \\
    & TERI + Decoder & 81.25 & 81.89 & 83.92 & 186.9 & 710.4 & 71.53 & 71.38 & 80.29 & 113.2 & 325.9 \\
    & TrajBERT + Decoder & 81.38 & 81.72 & 83.65 & 183.2 & 710.7 & 71.63 & 71.47 & 80.39 & 112.7 & 329.5 \\
    & RNTrajRec & 81.88 & 82.09 & 84.84 & 177.9 & 702.5 & 72.31 & 71.88 & 80.57 & 110.3 & 325.7\\
    & MM-STGED & 84.26 & 84.15 & 85.92 & 154.0 & 633.5 & 73.16 & 72.27 & 80.81 & 108.2 & 321.9\\
    \cmidrule{2-12}
    & \textbf{PLMTrajRec} & \underline{87.17} & \underline{87.99} & \underline{90.52} & \underline{141.4} & \underline{290.8} & \underline{74.42} & \underline{72.78} & \underline{82.82} & \underline{95.6} & \textbf{211.7}\\ 
    & \textbf{PLMTrajRec + FT} & \textbf{88.15} & \textbf{88.82} & \textbf{91.04} & \textbf{138.6} & \textbf{289.1} & \textbf{75.35} & \textbf{73.87} & \textbf{83.28} & \textbf{95.2} & \underline{220.3}\\
    \bottomrule
\end{tabular}
}
\label{tab:Porto}
\end{table*}

\subsection{Settings}
We split the trajectory dataset into training, validation, and testing sets in a 7:2:1 ratio. Following previous work~\cite{ren2021mtrajrec, chen2023rntrajrec, wei2024micro}, for dense map-matched trajectories, we set the sampling interval $\epsilon = 15$ seconds. To generate sparse trajectories, for each dense trajectory, we retain the first and last points and create three sparse versions by setting the sampling intervals $\mu = 4$ minutes, 2 minutes, and 1 minute, respectively. 

During training, we integrate sparse trajectories with three different sampling intervals to train PLMTrajRec. After training, we evaluate its performance in recovering trajectories across various sampling intervals to assess its generalizability.
To further enhance performance, we fine-tune the trained model for each specific sampling interval, allowing for a more precise recovery.

We employ PyTorch~\cite{paszke2019pytorch} framework to implement PLMTrajRec, with a learning rate of 1e-4 and a batch size of 64. BERT-small~\footnote{\url{https://github.com/google-research/bert}} is selected as the foundation model for PLM with 4 transformer layers and the number of hidden state is 512.
For road condition extraction, the area of interest is divided into a $64 \times 64$ grid, and the time dimension is partitioned into hourly intervals. The hidden state dimension is set to $F = 512$. In Equation~\ref{cal_road_weight}, we set $\kappa = 15$ and $\varphi_{dist} = 50$ meters. The model is trained for 50 epochs with early stopping, using a patience of 10 epochs. All experiments are conducted on NVIDIA RTX A4000 GPUs.

\begin{table*}[t]	
\small
\centering
\caption{Scalability analysis. The performance comparison on the Chengdu dataset when trained with different data ratios.}
\renewcommand{\arraystretch}{0.7}
\begin{tabular}{c|c|cc|cc|cc|cc|cc}
    \toprule
    \multirow{2}{*}{Sampling Interval}& Data Ratio &\multicolumn{2}{c|}{20\%} & \multicolumn{2}{c|}{40\%} & \multicolumn{2}{c|}{60\%} & \multicolumn{2}{c|}{80\%} & \multicolumn{2}{c}{100\%}\\
    \cline{2-12}
    & Metric & Acc($\%$) & RMSE & Acc($\%$) & RMSE & Acc($\%$) & RMSE & Acc($\%$) & RMSE & Acc($\%$) & RMSE \\
    \midrule
    \multirow{6}{*}{\parbox{2.5cm}{\centering $\mu = 2$ minutes $\to \epsilon = 15$ seconds}} & MTrajRec & 69.85 & 919.8 & 72.58 & 907.4 & 73.97 & 902.3 & 74.32 & 891.3 & 74.52 & 885.7 \\
    & T3s + Decoder & 68.86 & 909.5 & 72.29 & 897.2 & 73.63 & 883.8 & 74.53 & 868.8 & 74.62 & 857.5 \\
    & T2vec + Decoder & 69.47 & 905.4 & 72.77 & 858.3 & 73.80 & 855.8 & 74.89 & 831.3 & 75.69 & 783.6 \\
    & RNTrajRec & 68.05 & 991.0 & 70.20 & 856.9 & 71.17 & 834.4 & 72.62 & 795.3 & 75.80 & 757.0 \\
    & MM-STGED & 71.65 & 865.0 & 75.32 & 773.0 & 75.49 & 752.1 & 76.94 & 747.7 & 78.14 & 696.0\\
    & \textbf{PLMTrajRec} & \textbf{76.29} & \textbf{452.8} & \textbf{79.37} & \textbf{411.5} & \textbf{80.37} & \textbf{395.4} & \textbf{81.18} & \textbf{377.4} & \textbf{81.76} & \textbf{366.2} \\

    \bottomrule
\end{tabular}
\label{tab:scalibility}
\end{table*}

\subsection{Experimental Results}

Table~\ref{tab:Porto} reports a comparison of the results between our model and the baselines across various sampling intervals of sparse trajectories on both the Chengdu and Porto datasets. 
As the sampling interval increases, the map-matched trajectory recovery task is more difficult.
Due to the more complex road network structure in Porto compared to Chengdu, with Porto having 100.9 roads / $km^2$ and Chengdu having 38.1 roads / $km^2$, the accuracy of road segment recovery is lesser in Porto than in Chengdu.


Compared to other baselines, our model achieves superior performance across all metrics, with an average improvement of 16.51\% on the Chengdu dataset and 9.35\% on the Porto dataset. Notably, it shows substantial gains in RMSE metrics, with an average reduction of 351.8 meters in Chengdu and 119.2 meters in Porto. 
This indicates PLMTrajRec generalizes effectively, enabling it to accurately recover trajectories with varying sampling intervals.
By fine-tuning at the specific sampling interval, \textbf{PLMTrajRec + FT}, our model further improves Acc by 1.72\% and 0.32\%. 
These outstanding results can be attributed to the strong generalization capability of the PLM in processing sparse trajectories with varying sampling intervals. Additionally, the interval-aware trajectory embedder plays a pivotal role to capture spatial-temporal correlations and converting trajectory embeddings into a format that the PLM 
can effectively process and understand.


\subsection{Scalability Analysis}
To evaluate the effectiveness of PLMTrajRec in trajectory recovery when dense trajectory data is limited, we conduct scalability experiments on the Chengdu dataset.
Specifically, we train the model using subsets of the training set at 20\%, 40\%, 60\%, 80\%, and 100\%, and evaluate performance on the test set. 
The results are presented in Table~\ref{tab:scalibility} and \ref{tab:scalibility_14} of Appendix~\ref{Appdix_Sca}.
As the size of the training set increases, we observe performance improvements across all baseline models. Notably, with only 20\% of the training data, PLMTrajRec already surpasses most baselines, trailing only MM-STGED. With 40\% of the training data, PLMTrajRec outperforms all other models, demonstrating its strong scalability. This advantage underscores the practicality of PLMTrajRec in real-world scenarios, where obtaining large-scale dense trajectory data is often challenging.





\begin{table}[t!]	
\small
\centering
\caption{Zero-shot study on Chengdu dataset with sampling intervals at 2 minutes.}
\renewcommand{\arraystretch}{0.7}
\begin{tabular}{c|ccc}
    \toprule
    Methods & Acc($\%$) & MAE & RMSE \\
    \midrule
    MTrajRec & 56.39 ($\downarrow $ 18.13\%)& 492.7 ($\downarrow $ 48.37\%) & 1046.2 ($\downarrow $ 15.34\%)\\
    T3s + Decoder & 63.81 ($\downarrow $ 10.81\%) & 353.0 ($\downarrow $ 31.39\%)& 980.6 ($\downarrow $ 9.68\%)\\
    T2vec + Decoder & 63.32 ($\downarrow $ 12.37\%) & 376.6 ($\downarrow $ 38.50\%)& 928.1 ($\downarrow $ 15.57\%)\\
    RNTrajRec & 64.93 ($\downarrow $ 10.87\%) 
    & 339.6 ($\downarrow $ 35.70\%) & 912.8 ($\downarrow $ 17.07\%)\\
    MM-STGED & 68.27 ($\downarrow $ 9.90\%) & 291.8 ($\downarrow $ 32.42\%) & 862.6 ($\downarrow $ 19.31\%)\\
    \textbf{PLMTrajRec} & \textbf{79.62 ($\downarrow $ 2.14\%)} & \textbf{195.3 ($\downarrow $ 4.05\%)} & \textbf{379.6 ($\downarrow $ 3.53 \%)}\\
    \bottomrule
\end{tabular}
\label{tab:Generalization}
\end{table}
\subsection{Zero-shot Study on Sampling Interval}
We further analyze the generalizability of our model when the target sampling interval is absent during training. Specifically, PLMTrajRec is trained with a mixture of sampling intervals of 1 minute and 4 minutes and tested on a 2-minute sampling interval. The experimental results on the Chengdu dataset are shown in Table~\ref{tab:Generalization}, where $\downarrow$ indicates the percentage decline compared to training with a 2-minute sampling interval.
We observe that the accuracy of all models decreased when the target sampling interval was not included in the training data. However, our model shows a notably smaller performance drop. This robustness is attributed to the trajectory prompts and joint training strategy used by PLMTrajRec, which effectively identify the sampling interval of the target trajectories, demonstrating strong generalizability. In contrast, baseline models rely solely on trajectory point information to capture correlations, which struggles with generalization due to the inconsistency of temporal dynamics between the training and testing.

\subsection{Ablation Study}
To evaluate the effectiveness of the proposed components, we create six variants of PLMTrajRec. \textbf{w/o IF-guided explicit trajectory prompt}, \textbf{w/o AF-guided implicit trajectory prompt}, \textbf{w/o dual trajectory prompt}, \textbf{w/o road network}, \textbf{w/o reference tokens}, and \textbf{PLMTrajRec - Randomly initialized BERT}. The detailed description of each variant is shown in Appendix~\ref{des_variant}.

\begin{table}[t!]	
\small
\centering
\caption{Ablation study on Chengdu dataset with sampling intervals at 2 minutes.}
\renewcommand{\arraystretch}{0.9}
\resizebox{1.0\linewidth}{!}{
\begin{tabular}{c|ccccc}
    \toprule
    
    Methods & Acc($\%$) & Recall($\%$) & Prec($\%$) & MAE & RMSE \\
    \midrule
    w/o IF-guided explicit trajectory prompt & 81.04 & 84.08 & 88.07 & 205.8 & 394.2 \\
    w/o AF-guided implicit trajectory prompt & 81.43 & 84.28 & 88.14 & 191.4 & 371.1 \\
    w/o dual trajectory prompts & 80.70 & 83.61 & 87.73 & 225.9 & 407.4 \\
    w/o road network & 81.28 & 84.22 & 88.04 & 194.5 & 375.6 \\
    w/o reference tokens & 80.83 & 83.36 & 86.68 & 236.6 & 437.2 \\
    PLMTrajRec - Randomly initialized BERT & 74.11 & 78.62 & 80.83 & 288.4 & 831.9 \\
    \textbf{PLMTrajRec} & \textbf{81.76} & \textbf{84.31} & \textbf{88.38} & \textbf{187.4} & \textbf{366.2} \\
    
    \bottomrule
\end{tabular}
}
\label{tab:ablation}
\end{table}

As shown in Table~\ref{tab:ablation} and \ref{tab:ablation_porto}, each component of PLMTrajRec is essential. The IF-guided explicit trajectory prompt provides task and trajectory-related textual information, enabling PLMs to effectively understand trajectory features. Omitting it leads to a performance decrease. When the AF-guided implicit trajectory prompt is removed, many missing locations are replaced by the `[MASK]' token in BERT, limiting the model’s ability to capture road conditions for missing points, which results in suboptimal performance. 
For the trajectory point embedder, excluding road network for observed points causes the model to rely solely on latitude and longitude, neglecting the relationship between trajectories and the road network, which demonstrates that road network is crucial for modeling trajectory information. Additionally, removing reference tokens results in a performance decline, as PLMs are unable to directly interpret trajectory data. Using the randomly initialized BERT in PLMTrajRec has poor performance, indicating the efficiency of the PLM for trajectory recovery.

\section{Conclusion}
In this paper, we propose PLMTrajRec, a novel model for trajectory recovery leveraging PLM. PLMTrajRec effectively recovers trajectory even with limited dense trajectory data available, demonstrating strong scalability. The model equippes with an interval and feature-guided explicit trajectory prompt and interval-aware trajectory prompt enabling it to effectively generalize to different sampling intervals. 
Additionally, we introduce an area flow-guided implicit trajectory prompt to gather traffic flows in each region, and propose a road condition passing mechanism to infer missing-point conditions from nearby observations.
Experimental results on two datasets with three different sampling intervals validate the effectiveness, scalability, and generalizability of the proposed model.

\bibliographystyle{ACM-Reference-Format}
\bibliography{main}

\appendix


\section{Baseline Setting}
\label{append_baseline}
We choose the following 12 methods as baselines, including five free space trajectory recovery and seven map-matched trajectory recovery.
\subsection{Free-space Trajectory Recovery} 
Free space trajectory recovery first recovers trajectory points and then projects the trajectory onto the road network.
\begin{itemize}[leftmargin=*]
    \item \textbf{HMM~\cite{newson2009hidden} + ShortestPath} first projects the sparse trajectory on the road network based on the hidden markov model (HMM), and then calculate the shortest path.
    \item \textbf{Linear~\cite{hoteit2014estimating} + HMM~\cite{newson2009hidden}} linearly interpolates missing trajectory points and then implements HMM to perform the map matching process.
    \item \textbf{MPR~\cite{chen2011discovering} + HMM~\cite{newson2009hidden}} first divides the area of interest into grids to identify frequently traveled routes between sparse trajectory points. Then, it assumes that vehicle movement maintains a constant speed for trajectory recovery. Subsequently, it employs HMM to project the trajectory onto the road network.
    \item \textbf{DHTR~\cite{wang2019deep} + HMM~\cite{newson2009hidden}} incorporates a sequence-to-sequence framework and Kalman filtering to recover trajectory points. Subsequently, it applies an HMM to yield a trajectory that is constrained by the map.
    \item \textbf{AttnMove~\cite{xia2021attnmove} + Rule} uses attention to predict the missing road segments and uses the central location as the moving rate.
    
\end{itemize}
\subsection{Map-matched Trajectory Recovery}
Map-matched trajectory recovery can directly recover the trajectory on the road network.
\begin{itemize}[leftmargin=*]
    \item \textbf{MTrajRec~\cite{ren2021mtrajrec}} utilizes a sequence-to-sequence framework with Gated Recurrent Units (GRU) as the key component for trajectory recovery. It optimizes road segment and moving rate prediction through multi-task learning.
    \item \textbf{T2vec~\cite{li2018deep}} is a deep learning model for trajectory similarity learning. We use its encoder to embed the sparse trajectory.
    \item \textbf{T3s~\cite{yang2021t3s}} uses LSTM and attention mechanisms to encode sparse trajectory data effectively. 
    \item \textbf{TERI~\cite{chen2023teri}} assumes that the number of missing trajectory points is unknown and proposes a two-stage trajectory recovery framework. In the first stage, a transformer-based model predicts the number of points to be recovered, and in the second stage, the same framework is used for recovery trajectory coordinates. Here, we utilize only the second stage of TERI.
    \item \textbf{TrajBERT~\cite{si2023trajbert}}  employs a transformer encoder and a forward and backward neighbor selector to learn complex mobility patterns bi-directionally from sparse trajectories.
    \item \textbf{RNTrajRec~\cite{chen2023rntrajrec}} leverages the Transformer to capture the spatial-temporal correlation of the sparse trajectories. It also takes into account the relation between the trajectory and the road network.
    \item \textbf{MM-STGED~\cite{wei2024micro}} models sparse trajectories from a graph perspective and recovers trajectories by capturing micro and macro semantic information.
\end{itemize}

Notably, despite T2vec, T3S, TERI, and TrajBERT employing different techniques to capture sparse trajectories' spatial-temporal dependencies, they cannot directly generate the desired format for missing points. Therefore, after these models obtain the trajectory embeddings, we append the Decoder part of MTrajRec to output the road segment and moving ratio of the trajectory point. Denoted by \textbf{T2v + Decoder}, \textbf{T3S + Decoder}, \textbf{TERI + Decoder}, and \textbf{TrajBERT + Decoder}, respectively.

\section{Evaluation Metrics}
\label{append_metric}
We adopt five widely used metrics to evaluate the effectiveness of our model, following previous works~\cite{wei2024micro, chen2023rntrajrec, ren2021mtrajrec}. For road segment recovery, we use \textbf{Accuracy (Acc)}, \textbf{Recall}, and \textbf{Precision (Prec)} to assess the alignment between the true road segments $\mathcal{E}_p = \{e_1, \cdots, e_m\}$ and the predicted road segments $\hat{\mathcal{E}}_p = \{\hat{e}_1, \cdots, \hat{e}_m\}$. A higher value in these metrics indicates a more accurate road segment recovery. The metrics are formally defined as follows:
\begin{equation}
\begin{split}
    \textbf{Acc} &= \frac{1}{m}\sum_{i=1}^{m} \mathbbm{1} \{e_i=\hat{e}_i\}  \times 100\%,  \\
    \textbf{Recall} &= \frac{|\mathcal{E}_p \cap \hat{\mathcal{E} }_p  |}{|\mathcal{E}_p |} \times 100\% , \\
    \textbf{Prec} &= \frac{|\mathcal{E}_p \cap \hat{\mathcal{E} }_p  |}{|\hat{\mathcal{E}}_p |} \times 100\% ,
\end{split}
\end{equation}
where $\mathbbm{1} \{ \cdot \}$ is the indicator function, where $e_i=\hat{e}_i, \mathbbm{1} \{e_i=\hat{e}_i\} = 1$, elsewise  $\mathbbm{1} \{e_i=\hat{e}_i\} = 0$.

To evaluate the recovered GPS coordinates, we employ the \textbf{Mean Absolute Error (MAE)} and \textbf{Root Mean Square Error (RMSE)} to quantify the distance error between the true trajectory $\mathcal{T}_m = {q_1, \cdots, q_{|\mathcal{T}_m|}}$ and the predicted trajectory $\hat{\mathcal{T}}_m = {\hat{q}_1, \cdots, \hat{q}_{|\hat{\mathcal{T}}_m|}}$. The formulas for MAE and RMSE are given as follows:
The formulas of MAE and RMSE are as follows:
\begin{equation}
\begin{split}
    \textbf{MAE} &= \frac{1}{|\mathcal{T}_m| }\sum_{i=1}^{|\mathcal{T}_m|} |\mathrm{RN\_dist}(q_i, \hat{q}_i )| , \\
    \textbf{RMSE} &= \sqrt{\frac{1}{|\mathcal{T}_m| }\sum_{i=1}^{|\mathcal{T}_m|} |\mathrm{RN\_dist}(q_i, \hat{q}_i ) |^2}   
\end{split}
\end{equation}
Here, following~\cite{ren2021mtrajrec, chen2023rntrajrec, wei2024micro}, $\mathrm{RN\_dist} (q, \hat{q})$ signifies the shortest distance along the road network between the trajectory points $q$ and $\hat{q}$. Both MAE and RMSE are denoted in meters. Lower values of these metrics indicate a higher level of accuracy in the recovery results.

\section{Example of the IF-guided Explicit Trajectory Prompt}
\label{trajectory_prompt_example}
Consider a sparse trajectory $\mathcal{T} = \langle p_1, \cdots,p_N \rangle$ of $N$ trajectory points with a sampling interval of 4 minutes, starting at 8 o'clock on Saturday and ending at 9 o'clock on Saturday. Our goal is to recover it within a sampling interval of 15 seconds. 
Therefore, the trajectory prompts that are related to sampling intervals are: \textbf{Task Part}: Sparse trajectory recovery. \textbf{Target Part}: Output the road segment and moving ratio for each point in the trajectory. \textbf{Content Part}: The sparse trajectory is sampled every four minutes and aims to recover trajectory every fifteen seconds. The movement feature-related trajectory prompts are: \textbf{Time Part}: The trajectory started at eight o'clock on Saturday and ended at nine o'clock on Saturday. \textbf{Movement Part}: Total time cost: sixty minutes zero seconds. Total space transfer distance: $z$ kilometers. Here $z = {\textstyle \sum_{i=2}^{N}} \text{dist}(p_i, p_{i-1})$, where $\mathrm{dist}  (\cdot, \cdot)$ is used to calculate the distance between two trajectory points.

\section{Insight about the Learnable Fourier Features}
\label{insight_LLF}
Consider two trajectory points $x$ and $y$, and the feature mapping function $\Phi(x) = W_{\Phi } [\mathrm{cos} \text{ }x W_r || \mathrm{sin} \text{ }x W_r]$ in Learnable Fourier Features.
The relative information $x - y$ between points $x$ and $y$ can be captured through multiplication operations, i.e.:
\begin{equation}
    \begin{split}
\Phi(x)\cdot\Phi(y) &= W_{\Phi } [\mathrm{cos} \text{ }x W_r || \mathrm{sin} \text{ }x W_r] \cdot W_{\Phi } [\mathrm{cos} \text{ }y W_r || \mathrm{sin} \text{ }y W_r] \\
    &= ||W_{\phi}||_2 \cdot(\mathrm{cos}\text{ } x \cdot\mathrm{cos}\text{ } y + \mathrm{sin}\text{ } x \cdot\mathrm{sin}\text{ } y) \cdot||W_r||_2 \\
&= ||W_{\phi}||_2 \cdot\mathrm{cos}(x-y) \cdot||W_{r}||_2
\end{split}
\end{equation}

In our PLMTrajRec, after feature conversion, we input the trajectory feature into a pre-language training model based on BERT. Since there are a lot of multiplication-based attention operations in the PLM, the relative information $x - y$ can be easily modeled and utilized. This relative information helps infer crucial details like the distance between trajectory points, which is important for understanding movement. For instance, a larger distance between two points may indicate a higher likelihood of vehicle acceleration.

\section{The Detailed Description of Variant}
\label{des_variant}
\begin{itemize}
    \item \textbf{w/o IF-guided explicit trajectory prompt:} We remove the IF-guided explicit trajectory prompt. 
    \item \textbf{w/o AF-guided implicit trajectory prompt:} We use the `[MASK]' token in the BERT to represent the missing location.
    \item \textbf{w/o dual trajectory prompts:} We remove both the IF-guided explicit and implicit trajectory prompt, and the missing points are represented by the `[MASK]' token in BERT. 
    \item \textbf{w/o road network:} We only use the Learnable Fourier Feature for encoding the observed trajectory points while removing the road network feature. 
    \item \textbf{w/o reference tokens:} We remove the trajectory feature transformation layer in the trajectory embedder module.
    \item \textbf{PLMTrajRec - Randomly initialized BERT:} We randomly initialize the parameter of BERT instead of using the pre-trained BERT based on large-scale corpus datasets.
\end{itemize}
  
\section{Appendix Experiments}


\begin{table}[t]	
\centering
\caption{Hyperparameter range and optimal value.}
\begin{tabular}{c|c}
    \toprule
    Parameter & Range \\
    \midrule
    The number of reference tokens $K$ & 128, 256, \underline{512}, 1024\\
    LoRA rank $r$ & 4, \underline{8}, 16, 64 \\
    The loss function weight $\lambda$ & 0.1, 1, \underline{10}, 100\\
    \bottomrule
\end{tabular}
\label{tab:parameter}
\end{table}

\begin{figure}[h]
	\centering
	\subfigure{
		\includegraphics[width=0.48\linewidth]{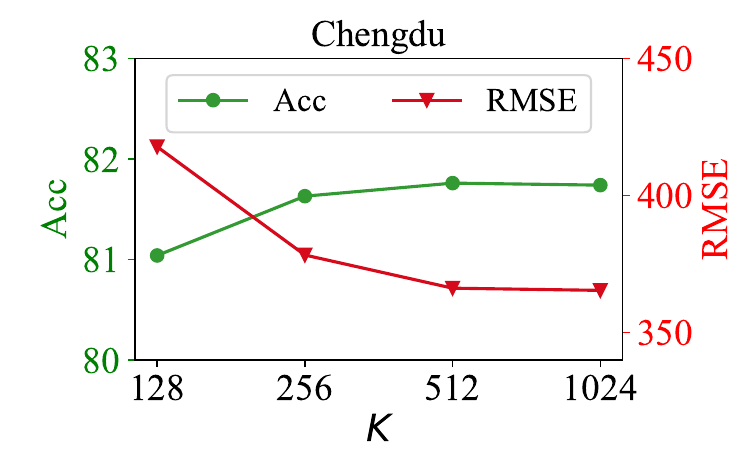}}
	\subfigure{
		\includegraphics[width=0.48\linewidth]{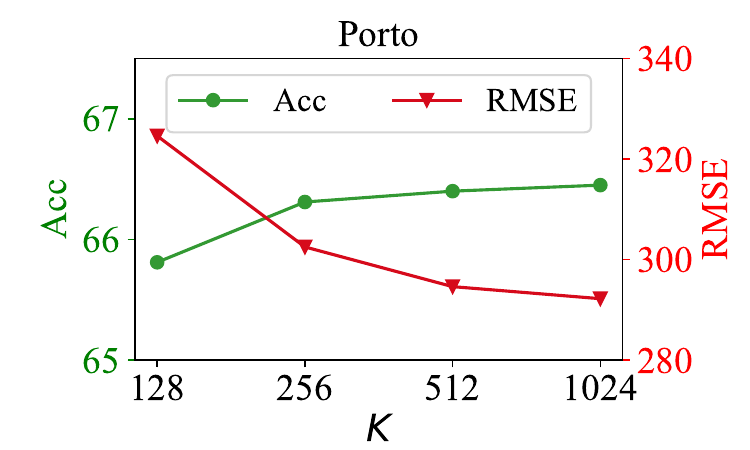}}
	\caption{Hyperparameter analysis of the number of reference tokens $K$ of trajectory feature transformation.}
	\label{fig:hypermeter_K}
\end{figure}

\begin{figure}[h]
	\centering
	\subfigure{
		\includegraphics[width=0.48\linewidth]{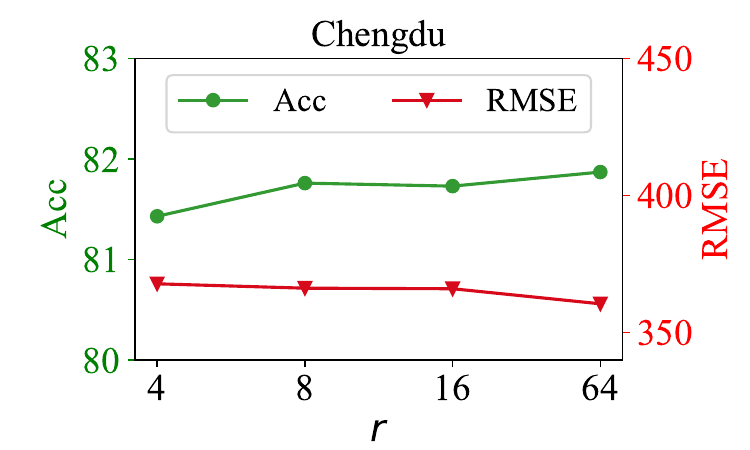}}
	\subfigure{
		\includegraphics[width=0.48\linewidth]{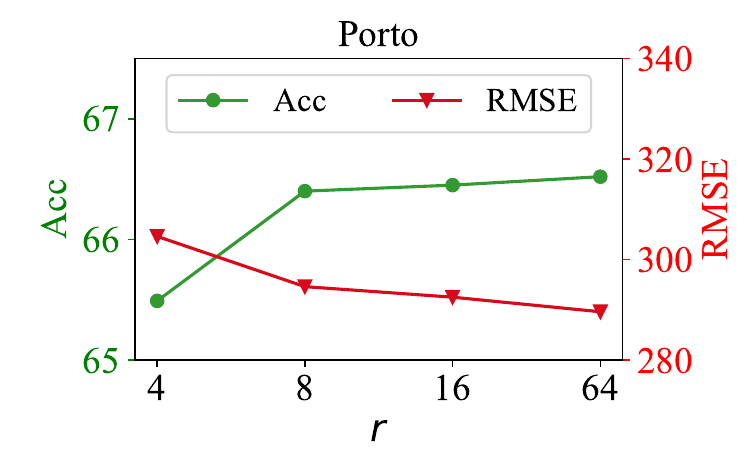}}
	\caption{Hyperparameter analysis of the rank $r$ of LoRA.}
	\label{fig:hypermeter_r}
\end{figure}

\begin{figure}[h]
	\centering
	\subfigure{
		\includegraphics[width=0.48\linewidth]{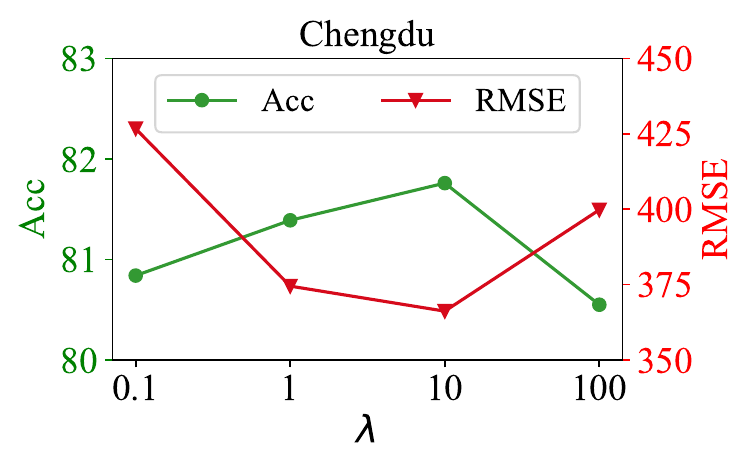}}
	\subfigure{
		\includegraphics[width=0.48\linewidth]{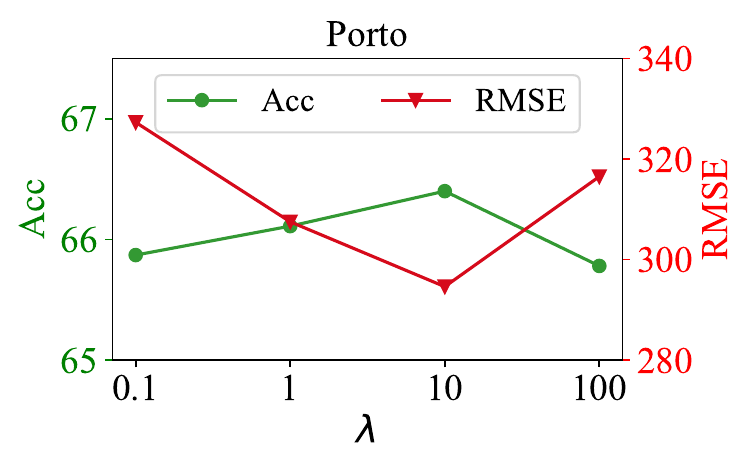}}
	\caption{Hyperparameter analysis of the weight $\lambda$ in the loss function.}
	\label{fig:hypermeter_lambda}
\end{figure}

\subsection{Hyperparameter Study}

To explore the impact of hyperparameters on model performance, we conduct hyperparameter analysis using the Chengdu and Porto datasets with the sparse trajectory sampling interval of 2 minutes. 
We consider three important hyperparameters, including the number of reference tokens $K$, the rank $r$ of LoRA, and the loss function weight $\lambda$. Details regarding the range of selection and the optimal values for these hyperparameters are presented in Table~\ref{tab:parameter}.


\begin{table*}[t!]	
\small
\centering
\caption{Scalability analysis on Chengdu dataset with 1 minute and 4 minutes sampling intervals.}
\renewcommand{\arraystretch}{0.7}
\begin{tabular}{c|c|cc|cc|cc|cc|cc}
    \toprule
    \multirow{2}{*}{Sampling Interval}& Data Ratio &\multicolumn{2}{c|}{20\%} & \multicolumn{2}{c|}{40\%} & \multicolumn{2}{c|}{60\%} & \multicolumn{2}{c|}{80\%} & \multicolumn{2}{c}{100\%}\\
    \cline{2-12}
    & Metric & Acc($\%$) & RMSE & Acc($\%$) & RMSE & Acc($\%$) & RMSE & Acc($\%$) & RMSE & Acc($\%$) & RMSE \\
    \midrule
    \multirow{6}{*}{\parbox{2.5cm}{\centering $\mu = 4$ minutes $\to \epsilon = 15$ seconds}} & MTrajRec & 60.73 & 1048.0 & 63.58 & 968.5 & 64.73 & 929.4 & 65.39 & 914.6 & 65.79 & 904.4 \\
    & T3s + Decoder & 60.49 & 1098.9 & 63.32 & 968.7 & 64.23 & 961.9 & 65.04 & 942.3 & 65.60 & 926.3 \\
    & T2vec + Decoder & 62.17 & 966.1 & 64.92 & 946.1 & 64.95 & 972.3 & 65.31 & 946.8 & 66.51 & 915.2 \\
    & RNTrajRec & 60.80 & 998.0 & 62.85 & 931.3 & 64.19 & 909.1 & 65.46 & 835.0 & 67.66 & 886.0 \\
    & MM-STGED & 64.61 & 935.0 & 67.76 & 881.8 & 68.53 & 853.7 & 69.95 & 825.3 & 70.64 & 829.7\\
    & \textbf{PLMTrajRec} & \textbf{68.30} & \textbf{585.3} & \textbf{71.57} & \textbf{515.6} & \textbf{72.74} & \textbf{512.5} & \textbf{73.56} & \textbf{488.3} & \textbf{74.12} & \textbf{483.0} \\
    
    \midrule
    \multirow{6}{*}{\parbox{2.5cm}{\centering $\mu = 1$ minute $\to \epsilon = 15$ seconds}} & MTrajRec & 75.39 & 937.5 & 78.53 & 835.1 & 80.01 & 794.3 & 80.93 & 725.7 & 81.12 & 718.4 \\
    & T3s + Decoder & 76.49 & 917.1 & 79.08 & 824.9 & 80.65 & 767.8 & 80.82 & 716.5 & 80.90 & 713.0 \\
    & T2vec + Decoder & 75.89 & 845.4 & 79.09 & 746.8 & 79.41 & 752.3 & 81.21 & 742.2 & 81.69 & 714.1 \\
    & RNTrajRec & 75.65 & 846.2 & 79.48 & 784.8 & 79.61 & 769.3 & 81.74 & 742.5 & 81.88 & 702.5\\
    & MM-STGED & 76.02 & 857.7 & 79.86 & 734.7 & 82.25 & 676.4 & 83.44 & 663.0 & 84.26 & 633.5 \\
    & \textbf{PLMTrajRec} & \textbf{81.62} & \textbf{397.5} & \textbf{84.59} & \textbf{352.0} & \textbf{85.75} & \textbf{328.7} & \textbf{86.73} & \textbf{305.4} & \textbf{87.17} & \textbf{290.8} \\
    \bottomrule
\end{tabular}
\label{tab:scalibility_14}
\end{table*}

\subsubsection{The Number of Reference Tokens $K$} 
We set the parameter $K$ from the set $\{128, 256, 512, 1024\}$ to explore its impact on trajectory recovery. The experimental results are shown in Figure~\ref{fig:hypermeter_K}. As $K$ increases, the effectiveness of trajectory recovery also improves. This suggests that having a larger token space aids in accurately representing trajectory features.
When $K$ is larger than 512, the performance improvement of PLMTrajRec becomes marginal, while the computational cost will increase. To balance the performance and efficiency of the model, we set $K$ to 512.

\subsubsection{The Rank $r$ of LoRA}
As shown in Figure~\ref{fig:hypermeter_r}, we set the range of $r$ to 4, 8, 16, and 64 to explore its impact. It is observed that the model performs well when $r = 4$. 
As $r$ increases, the accuracy will continue to increase, but it is not obvious. This suggests that PLMs encapsulate substantial domain expertise through training on extensive corpora, rendering them adaptable to trajectory recovery tasks with minor adjustments. Yet as $r$ increases, the number of parameters also increases, making the model training require more memory. Therefore, we set $r = 8$ to balance the performance and resources.

\subsubsection{The Loss Weight $\lambda$}
As shown in Figure~\ref{fig:hypermeter_lambda}, as the loss function weight $\lambda$ increases, the model performance first improves and then decreases. This is because a smaller $\lambda$ will make the model focus on road segment recovery, while a larger $\lambda$ will focus on moving rate recovery. To balance these two tasks, we set the value of $\lambda$ to 10.

\subsection{Scalability Study on 1-minute and 4-minute Sampling Interval}
\label{Appdix_Sca}
The result of the scalability study on the sparse trajectory with 1-minute and 4 minutes, as shown in Table~\ref{tab:scalibility_14}, the conclusion is consistent with the 2-minute sampling interval. The experiment results prove the scalability of PLMTrajRec.

\begin{table}[t]	
\small
\centering
\caption{Ablation study on Porto dataset with sampling intervals at 2 minutes.}
\renewcommand{\arraystretch}{0.7}
\resizebox{1.0\linewidth}{!}{
\begin{tabular}{c|ccccc}
    \toprule
    
    Methods & Acc($\%$) & Recall($\%$) & Prec($\%$) & MAE & RMSE \\
    
    \midrule
    w/o IF-guided explicit trajectory prompt & 66.13 & 66.19 & 81.98 & 157.9 & 314.6 \\
    w/o AF-guided implicit trajectory prompt & 66.18 & 66.22 & 82.01 & 149.6 & 298.1 \\
    w/o dual trajectory prompts & 65.52 & 65.89 & 81.51 & 171.0 & 339.4 \\
    w/o road network & 66.11 & 66.27 & 81.90 & 150.7 & 302.8 \\
    w/o reference tokens & 65.89 & 65.07 & 80.57 & 165.4 & 327.7 \\
    PLMTrajRec - Randomly initialized BERT & 62.06 & 64.18 & 79.26 & 179.3 & 413.9\\
    \textbf{PLMTrajRec}  & \textbf{66.40} & \textbf{66.52} & \textbf{82.14} & \textbf{141.9} & \textbf{294.6} \\
    \bottomrule
\end{tabular}
}
\label{tab:ablation_porto}
\end{table}
\subsection{Ablation Study on the Porto}
The result of the ablation study on the Porto dataset, as shown in Table~\ref{tab:ablation_porto}, the conclusion is consistent with the Chengdu dataset. Each component contributes to PLMTrajRec's performance, and removing any component results in a decline in model effectiveness.


\begin{table}[t]	
	\centering
	\caption{Computation Cost.}
	\renewcommand{\arraystretch}{0.8}
	
	\begin{tabular}{c|c|c|c}
		\toprule
            Dataset & \multicolumn{3}{c}{Chengdu / Porto} \\
            \midrule
		\multirow{2}{*}{Methods} & Model size & Train time & Inference time \\
            & (MBytes) & (min/epoch) & (min) \\
		\midrule
		DHTR & 19 / 18 & 11.30 / 42.55 & 1.13 / 2.35\\
		MTrajRec & 49 / 44 & 37.19 / 142.12 & 12.57 / 28.20 \\ 
		RNTrajRec & 112 / 109 & 72.30 / 153.29 & 29.14 / 43.58 \\
		MM-STGED & 34 / 31 & 41.27 / 138.33 & 16.27 / 30.58 \\ 
            \textbf{PLMTrajRec} & \textbf{176 / 175} & \textbf{24.18 / 54.14} & \textbf{2.56 / 10.38}\\
		\bottomrule
	\end{tabular}
	\label{tab:cost}
\end{table}

\subsection{Computational Cost}
We use an NVIDIA RTX A4000 card to conduct cost analysis on the Chengdu and Porto datasets with a 2-minute sampling interval by considering model size, training time, and inference time.
As shown in Table~\ref{tab:cost}, compared to the end-to-end deep learning model, PLMTrajRec has the largest number of parameters due to its utilization of a pre-trained language model based on the BERT-small framework.
Despite having a greater number of parameters, PLMTrajRec exhibits quicker training and testing times as it can output results for all trajectory points at once, and the majority of parameters are frozen, unlike other models that necessitate autoregressive generation. It significantly accelerates both model training and inference processes.

	
	

\begin{figure}[h]
  \centering
  \includegraphics[width=\linewidth]{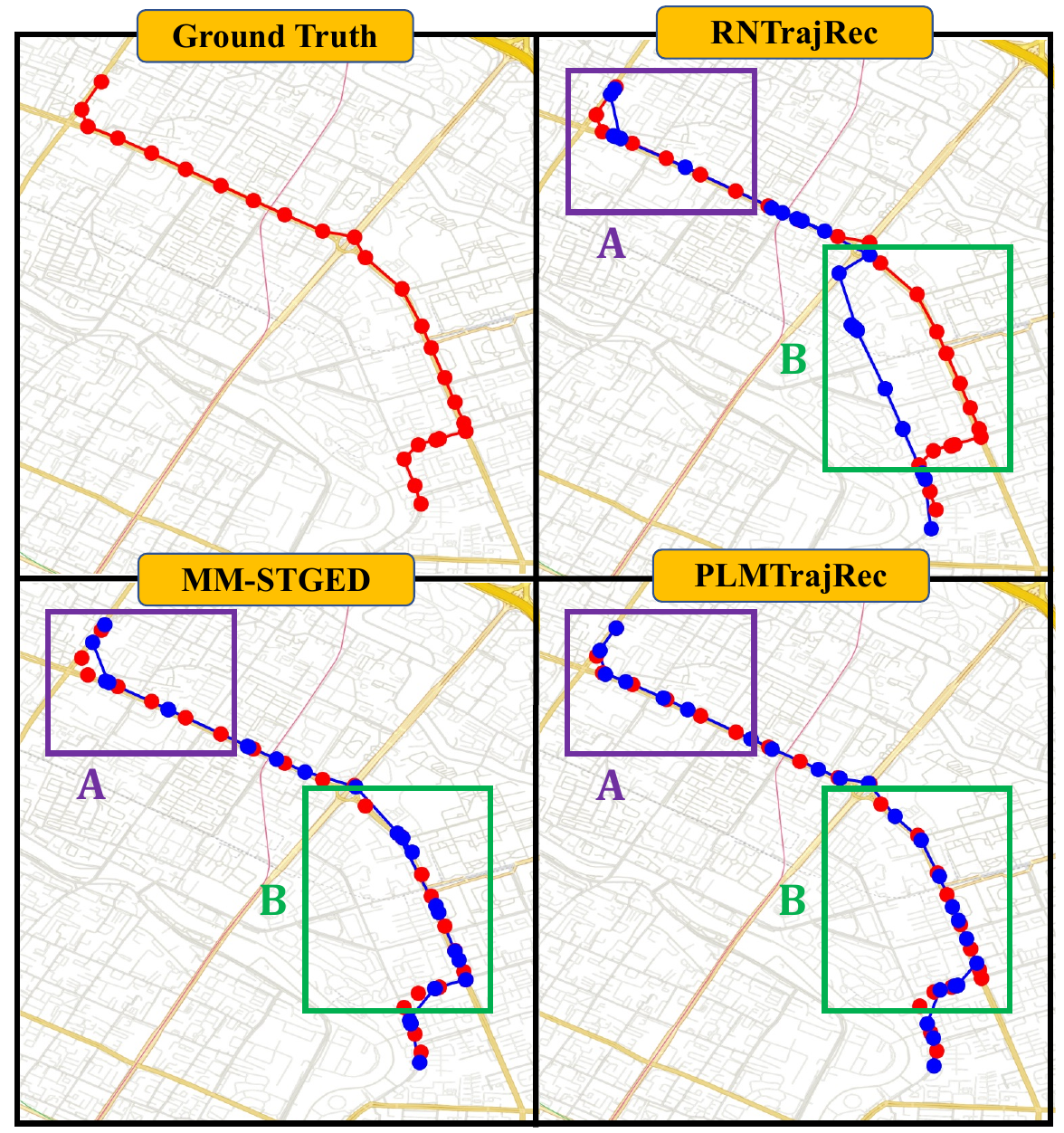}
  \caption{Case study on the Chengdu dataset. Red points represent the truth trajectory points and blue points represent the recovered trajectory points.}
  \vspace{0.01cm}
  \label{fig:case_CD}
\end{figure}

\subsection{Case Study}
We conduct a case study on the Chengdu dataset to visualize the trajectory recovery performance with various baselines.
As shown in Figure~\ref{fig:case_CD}, we draw the truth and recovered trajectory, where red points represent the truth trajectory points and blue points indicate the recovered trajectory points.
We find that the recovered trajectory aligns well with the road network, demonstrating the effectiveness of using road segment and moving rate to represent trajectory point. 
To facilitate a more intuitive comparison of recovery performance among different models, we focus on two distinct regions, labeled as A and B, for visual analysis. 
In region A, characterized by a relatively simple road network structure, all models can accurately recover road segments. Among them, PLMTrajRec maps trajectory points better by leveraging PLMs. 
In contrast, region B exhibits a more intricate road network with multiple accessible routes.
RNTrajRec captures spatial-temporal correlations of trajectories that are not adequate and recovers incorrect road segments. 
PLMTrajRec not only excels in road segment recovery but also in accurately matching the actual trajectory points. 
This success can be attributed to its ability to model missing trajectory points through the implicit trajectory prompt, thereby introducing more valuable information and improving performance.

\end{document}